\documentclass[12pt]{article}
\pdfoutput=1
\usepackage{graphicx,pict2e,latexsym,xcolor}
\usepackage[square,numbers]{natbib}
\newtheorem{defn}{Definition}
\newtheorem{theorem}{Theorem}
\newtheorem{lemma}{Lemma}
\newtheorem{cor}{Corollary}

\def\E{\mathbf E}

\def\G{\mathcal G}

\def\P{\mathbf P}

\def\M{\mathcal M}

\def\B{\mathcal B}

\newcommand{\iid}{\stackrel{iid}{\sim}}

\begin{document}
\bibliographystyle{plainnat}
\pagestyle{plain}

\title{\Large \bf Learning Submanifolds  
for Subsequent Inference \\ on Random Dot Product Graphs, \\
Part 1: Theory}
\author{Michael W. Trosset\thanks{Department of Statistics, Indiana University.  
E-mail: {\tt mtrosset@iu.edu}} 
\and
Carey E. Priebe\thanks{Department of Applied Mathematics \& Statistics, Johns Hopkins University. 
E-mail: {\tt cep@jhu.edu}}
}

\date{\today}

\maketitle


\begin{abstract}
We propose a framework for restricted inference on random dot product graphs whose latent positions lie on an unknown low-dimensional support manifold.  For general decision problems, we propose semisupervised decision rules that use auxiliary data to learn the support manifold.  Specifically, our rules use the Isomap manifold learning procedure to construct a low-dimensional Euclidean representation of the observed graph, in which space an isometrically invariant function maps configurations of points to actions.  We study the behavior of the proposed rules as the quantity of auxiliary data sampled from the unknown support manifold increases.  We show that, as the auxiliary sample size increases, the risk of the semisupervised rule converges to the risk of an oracle rule that relies on the maximal amount of low-dimensional Euclidean structure that can be extracted from the support manifold.  Examples, applications, and simulation studies are deferred to a sequel.
\end{abstract}


\newpage

\tableofcontents

\newpage


\section{Introduction}
\label{intro}

Networks are complex data structures that describe connections between experimental units (objects, subjects, individuals).  Mathematically, networks are graphs.  The units are modeled as vertices, the connections as edges.  Networks arise in an enormous variety of disciplines, including networks of neural pathways (connectomes) in brain science, networks of genes that interact to control cell functions (gene regulatory networks) in biology, and networks of social interactions (social networks) in sociology, political science, etc.
Network science studies the structure of such networks.

Network machine learning is an important component of network science.
To perform statistical inference using network data, it is necessary to specify a probability model that is assumed to have generated the observed network(s).  Such models are called random graphs.  The random graphs of interest here assume a fixed set of vertices, with edges formed by performing independent Bernoulli trials on the possible pairs of vertices.
An early example is the Erd\"{o}s-R\'{e}nyi model proposed by Gilbert \cite{Gilbert:1959}, in which each possible edge occurs with fixed probability $p$.  This requirement is too strong for most applications; various researchers have weakened it in various ways.

Stochastic blockmodels \cite{holland&etal:1983,anderson&etal:1992} generalize the Erd\"{o}s-R\'{e}nyi model; in turn, random dot product graphs (RDPGs, \cite{RDPG:2007}) generalize stochastic blockmodels. 
Random dot product graphs also exemplify the latent space approach to network analysis considered in \cite{hoff&etal:2002}. 
There now exists a considerable body of theory and methodology for statistical inference on RDPGs; see, for example, \cite{AA&etal:2018} and the references therein.

Athreya and collaborators \cite{LSM:2018} studied the special case of an RDPG whose latent positions lie on a support curve.  This is the case that concerned us in \cite{mwt:rdpg1}, except that we supposed the curve to be completely unknown and attempted to learn it well enough to exploit its structure in certain $1$- and $2$-sample tests.  The present work extends our investigation to the case of an RDPG whose latent positions lie on a low-dimensional support surface.
As in \cite{mwt:rdpg1}, we submit that procedures that exploit the surface's structure should be more effective than procedures that do not.  The challenge is how to exploit that structure when it is unknown.  In further contrast to \cite{mwt:rdpg1}, we consider a general class of statistical decision problems on the RDPG.  The methods that we propose can be extended to the case of generalized RDPGs, as defined in \cite{GRDPG:2022}, but the notation is more cumbersome.

We provide a succinct explication of RDPGs in Section \ref{rdpg} and propose a class of semisupervised statistical decision problems in Section \ref{infer}.  Our convergence analysis is presented in Section \ref{converge}.  Here we demonstrate that, as more and more auxiliary information about the underlying support surface is obtained, the performance of our semisupervised decision rule converges to the performance of an oracle decision rule that fully exploits the low-dimensional Euclidean structure of the support surface.  Section \ref{discuss} concludes Part~1.  Examples, applications, and simulation studies will be reported in Part~2.

\section{Random Dot Product Graphs}
\label{rdpg}

A {\em graph}\/ is an ordered pair ${\mathcal G}=(V,E)$, where $V=\{1,\ldots,k\}$ is the {\em vertex set}\/ and $E \subset V \times V$ is the {\em edge set}.  There is an edge between vertices $i$ and $j$ if and only if $(i,j) \in E$.  Alternatively, the $k \times k$ {\em adjacency matrix}\/ $A$ of ${\mathcal G}$ is the 0-1 matrix defined by $A_{ij}=1$ if and only if $(i,j) \in E$.
The graph ${\mathcal G}$ is undirected if $A$ is symmetric and simple if $A$ is hollow, i.e., if each $A_{ii}=0$.  A {\em random graph}\/ is a probability model for generating simple undirected graphs, or (equivalently) their adjacency matrices.

Latent position graphs were proposed in \cite{hoff&etal:2002}.
In these models, each vertex is associated with a position in a latent (unobserved) space and edge probabilities are determined by a {\em link function}\/ of pairs of positions.  
\begin{defn}[conditional latent position graph]
Fix $\M \subseteq \Re^q$ and let $f : \M \times \M \rightarrow [0,1]$ be a symmetric link function.
Fix latent positions $x_1,\ldots,x_k \in \M$, then form the $k \times q$ latent position matrix
\[
X = \left[ \begin{array}{c|c|c} x_1 & \cdots & x_k \end{array} \right]^\top.
\]
Suppose that $A$ is a symmetric hollow adjacency matrix whose above-diagonal entries are independent Bernoulli trials with success probabilities $P(A_{ij}=1)= f(x_i,x_j)$.  We then write $A \sim \mbox{\tt LPG}_f(X)$ and say that $A$ is the adjacency matrix of a conditional latent position graph with latent positions $x_1,\ldots,x_k$.
\label{def:CLPG}
\end{defn}
In many applications, one assumes that the latent positions $x_1,\ldots,x_k \iid P$, a probability distribution on $\M$,
in which case we write $(A,X) \sim \mbox{\tt LPG}_f(P)$.  It is with this understanding that the LPGs in Definition~\ref{def:CLPG} are conditional.
Popular link functions include
$f(x_i,x_j) = \exp(-\|x_i-x_j\|^2/(2\sigma^2))$ and
$f(x_i,x_j) = x_i^\top x_j$.  The latter defines the {\em random dot product graph}\/ (RDPG) \cite{RDPG:2007}.  The positive definite Euclidean inner product in the RDPG model was replaced by an indefinite inner product in \cite{GRDPG:2022}, resulting in the {\em generalized}\/ RDPG (GRDPG).

Recent work by Rubin-Delanchy and collaborators
\cite{PRD:2020,WhiteleyGrayPRD:2021,WhiteleyGrayPRD:2022}
has revealed a remarkable low-dimensional manifold structure common to ``well-behaved'' LPGs.  
Following \cite{PRD:2020},
assume that $f \in L^2(\Re^q \times \Re^q)$ and define ${\mathcal A} : L^2(\Re^q) \rightarrow \Re$ by
\[
{\mathcal A}g(x) = \int_{\Re^q} f(x,y) g(y) \, dy.
\]
Let $\lambda_1 \geq \lambda_2 \geq \ldots$ denote the nonzero eigenvalues of ${\mathcal A}$, with corresponding orthonormal eigenfunctions $u_1, u_2,\ldots \in L^2(\Re^q)$.  Assuming that ${\mathcal A}$ is trace-class, i.e., the $|\lambda_j|$ are summable,
define the kernel function $\kappa \in L^2(\Re^q \times \Re^q)$ by
\begin{eqnarray*}
\kappa(x,y) = \sum_j \left| \lambda_j \right|^{1/2} u_j(x) u_j(y).
& &
\mbox{Setting }
[g,h] = \sum_j \mbox{sgn} \left( \lambda_j \right)
\langle g,u_j \rangle \, \langle h,u_j \rangle,
\end{eqnarray*}
it is easily shown that the indefinite inner product $[\kappa(x,\cdot),\kappa(y,\cdot)] = f(x,y)$ almost everywhere.
Hence, the mapping $\phi(X) = \kappa(X,\cdot)$ embeds latent positions in $L^2(\Re^q)$, in which space the edge probabilities are specified by a GRDPG.

To estimate $\phi(x_1),\ldots,\phi(x_k)$, one approximates the operator ${\mathcal A}$ with the observed adjacency matrix $A$.
\begin{defn}[generalized adjacency spectral embedding]
Consider the eigenvalues of adjacency matrix $A$ and let
$\hat{\lambda}_1,\cdots,\hat{\lambda}_r$ denote the $r$ eigenvalues of largest absolute values.  Let $\hat{u}_1,\ldots,\hat{u}_r$ denote corresponding eigenvectors and set $\sigma_i^2 = |\lambda_i|$.
The {\em adjacency spectral embedding}\/ (ASE) of $A$ in $\Re^r$ is
$\hat{X} = U_r {S}_r$, where $U_r$ is the $k \times r$ matrix whose columns are $\hat{u}_1,\ldots,\hat{u}_r$ and ${S}_r$ is the $r \times r$ diagonal matrix whose diagonal entries are $\sigma_1,\ldots,\sigma_r$.
\label{def:ASE}
\end{defn}

Now assume that $x_1,\ldots,x_k \iid F$, where $F$ is absolutely continuous with respect to Lebesgue measure on $\M$. 
Assuming further that the absolute value of the operator ${\mathcal A}$ is H\"{o}lder continuous with exponent $\alpha$, Rubin-Delanchy \cite{PRD:2020} bounded the Hausdorff dimension of $\phi(\M) \subset L^2(\Re^q)$ by $2q/\alpha$.
Stronger assumptions in \cite{WhiteleyGrayPRD:2022} ensure that,
if $\M \subset \Re^q$ is a Riemannian manifold, then $\phi(\M) \subset L^2(\Re^q)$ is a Riemannian manifold that is diffeomorphic to $\M$.
Thus, every sufficiently well-behaved LPG is equivalent to a GRDPG with latent positions on a low-dimensional submanifold of $L^2(\Re^q)$.  

Motivated by this insight, we have been developing theory and methods for learning and exploiting the structure of submanifolds of latent positions in RDPGs and GRDPGs.
A first step was taken by Atheyra et al.\ \cite{LSM:2018}, who studied the special case of an RDPG whose latent positions lie on a ``structural support curve.''  
In \cite{mwt:rdpg1}, we supposed the curve to be completely unknown and proposed methods that learn it well enough to exploit its structure in subsequent inference.
The intent of the proposed project is to extend these methods and the accompanying theory from $1$-dimensional curves to $p$-dimensional surfaces.  The extension is nontrivial because the crucial steps in our $1$-dimensional analysis relied on the fact that all $1$-dimensional submanifolds are Euclidean, i.e., isometric to a subset of Euclidean space.  Although multidimensional Euclidean surfaces do exist (e.g., Swiss rolls), multidimensional surfaces are generally not isometric to subsets of Euclidean space.  New approaches were needed to address the multidimensional setting.

Specializing Definition~\ref{def:CLPG} to the link function $f(x_i,x_j) = x_i^\top x_j$, we obtain
\begin{defn}[conditional RDPG]
Fix $x_1,\ldots,x_k \in \Re^q$, then form the $k \times q$ latent position matrix
\[
X = \left[ \begin{array}{c|c|c} x_1 & \cdots & x_k \end{array} \right]^\top.
\]
Suppose that $A$ is a symmetric hollow adjacency matrix whose above-diagonal entries are independent Bernoulli trials with success probabilities $P(A_{ij}=1)=x_i^\top x_j$.  We then write $A \sim \mbox{\tt RDPG}(X)$ and say that $A$ is the adjacency matrix of a random dot product graph with latent positions $x_1,\ldots,x_k$.
\label{def:CRDPG}
\end{defn}

We are concerned with drawing inferences about RDPGs when the latent positions are restricted to lie on a $p$-dimensional Riemannian manifold ${\mathcal M} \subset \Re^q$.  We require the concept of an inner product distribution in order to sample $\M$, which in turn allows us to define an RDPG with random latent positions.
\begin{defn}[inner product distribution]
A probability distribution $P$ with support ${\mathcal M} \subset \Re^q$ is a $q$-dimensional inner product distribution if and only if $x,y \in {\mathcal M}$ entails $x^\top y \in [0,1]$.
\end{defn}
Notice that, for $P$ to be an inner product distribution, its support must lie in the nonnegative orthant of the $q$-dimensional unit ball; hence, the extrinsic diameter of $\M$ is bounded above by $\sqrt{2}$.
\begin{defn}[RDPG with random latent positions]
Suppose that $x_1,\ldots,x_n$ are random vectors in $\M$ with joint distribution $J$
If $A|X \sim \mbox{\tt RDPG}(X)$, then we write $(A,X) \sim \mbox{\tt RDPG}(J)$ and say that $A$ is the adjacency matrix of a random dot product graph with random latent positions $x_1,\ldots,x_k$.
In the special case that $P$ is an inner product distribution and that $x_1,\ldots,x_k \iid P$, then we simply write $(A,X) \sim \mbox{\tt RDPG}(P)$.
 \label{def:iidRDPG}
\end{defn}
Taking $J$ to represent point-mass at $(x_1,\ldots,x_n) \in \M^n$, we see that Definition~\ref{def:CRDPG} can be viewed as a special case of Definition~\ref{def:iidRDPG}.

\section{Restricted Inference on RDPGs}
\label{infer}

Suppose that the possible latent position vectors for an RDPG lie on a connected compact Riemannian manifold $\M \subset \Re^q$.
A function $\Delta : \M \times \M \rightarrow [0,\infty)$ is a {\em dissimilarity function}\/ on $\M \times \M$ if and only if $\Delta(x,x) = 0$ for every $x \in \M$ and $\Delta(x,y)=\Delta(y,x)$ for every $(x,y) \in \M \times \M$.  We denote the Riemannian distance function on $\M \times \M$ by $\Delta_\infty$.

Let $\B$ denote the sigma-field generated by the open balls of $\Delta_\infty$.  We will denote probability measures on $(\M,\B)$ by $P : \B \rightarrow [0,1]$.  The experiment of interest is $(A,L) \sim \mbox{\tt RDPG}(J)$, i.e., the adjacency matrix $A$ is generated from latent positions $\ell_1,\ldots,\ell_n$ drawn from the joint probability distribution $J$.  For example, a $2$-sample problem might entail drawing $\ell_1,\ldots,\ell_{n_1} \iid P_1$ and $\ell_{n_1+1},\ldots,\ell_{n_1+n_2} \iid P_2$.

Adopting the language of statistical decision theory,
let $\Theta$ denote the possible states of nature, 
let \mbox{\tt Act} denote the space of possible actions, and let
$\mbox{\tt Loss}(\theta,a)$ denote the loss that is incurred if $\theta \in \Theta$ obtains and one takes action $a \in \mbox{\tt Act}$.
We will study the behavior of randomized decision rules of the following form:
\begin{enumerate}

\item Let $P_\infty$ denote a probability measure on $(\M,\B)$ with full support, i.e., that assigns strictly positive probability to each open ball.  Draw auxiliary latent positions $m_1,\ldots,m_N \iid P_\infty$ and let $\hat{P}_N$ denote their empirical distribution.

\item Generate the augmented adjacency matrix $A_N^+ =
\left[ \begin{array}{c|c} 
A & \cdot \\ \hline \cdot & \cdot
\end{array} \right]$.

\item Estimate the latent positions as
$\hat{\ell}_1,\ldots,\hat{\ell}_n,\hat{m}_1,\ldots,\hat{m}_N \in \Re^q$ by ASE.

\item Construct an $(n+N) \times (n+N)$ dissimilarity matrix $[ \Delta_N(x,y) ]$ from the estimated latent positions by computing shortest path distances on a suitable graph.

\item Embed the relevant dissimilarities in $\Re^d$ by minimizing Kruskal's \cite{kruskal:1964a} raw stress criterion.
We consider three strategies.  
The {\em partial graph embedding}\/ (PGE) strategy constructs $\bar{\ell}_1,\ldots,\bar{\ell}_n \in \Re^d$ from the $n \times n$ submatrix $[ \Delta_N(\hat{\ell}_r,\hat{\ell}_s) ]$.
The {\em complete graph embedding}\/ (CGE) strategy constructs
$\bar{\ell}_1,\ldots,\bar{\ell}_n,\bar{m}_1,\ldots,\bar{m}_N \in \Re^d$ from $[ \Delta_N(x,y) ]$.  CGE with Approximate Lipschitz Embedding (ALE) also uses $[ \Delta_N(x,y) ]$, but imposes smoothing restrictions on the configuration.

\item  Take action $\phi(\bar{\ell}_1,\ldots,\bar{\ell}_n) = a \in \mbox{\tt Act}$, where $\phi : \Re^{n \times d} \rightarrow \mbox{\tt Act}$ is invariant under isometric transformations.

\end{enumerate}
Such rules are semisupervised inferences in the sense that the 
auxiliary latent positions are used to estimate $\ell_1,\ldots,\ell_n$ (from $A^+$ rather than from $A$); to learn $\M$, by which we mean to approximate Riemannian distance on $\M$ with shortest path distance on a suitable graph; and to construct the $d$-dimensional Euclidean representation of the estimated latent positions.  After this representation has been constructed, however, the action function $\phi$ depends entirely on $\bar{\ell}_1,\ldots,\bar{\ell}_n \in \Re^d$.

We study the behavior of the proposed rules as the amount of auxiliary information about $\M$ increases.  We show that, as $N \rightarrow \infty$, the performance of the semisupervised rule converges to the performance of an oracle rule that relies on the maximal amount of Euclidean structure that can be extracted from $\M$.  We emphasize that this finding is {\em not}\/ a consistency result.  Because we hold $n$ fixed, the performance of {\em any}\/ rule will be poor if $n$ is small.
Rather, we argue that, if $\M$ is such that inference restricted to $\M$ is beneficial, then it is possible to learn $\M$ well enough to accrue the benefits of restricted inference.

Because $\phi$ is invariant under isometric transformation, we can rewrite $\phi(\bar{\ell}_1,\ldots,\bar{\ell}_n)$ as $\phi(E)$, where $E$ is the $n \times n$ matrix of interpoint Euclidean distances $\| \bar{\ell}_r-\bar{\ell}_s \|$.  The random matrix $E$ depends on the original latent positions and the auxiliary latent positions through the adjacency matrix $A^+$ generated by the RDPG model.  The maximal information that we might extract from $m_1,m_2,\ldots \iid P_\infty$ is the exact latent positions $\ell_1,\ldots,\ell_n$, the exact Riemannian distance function $\Delta_\infty : \M \times \M \rightarrow \Re$, and the exact auxiliary sampling distribution $P_\infty$.  In this case, we write $E = E_\infty ( \ell_1,\ldots,\ell_n;\Delta_\infty,P_\infty)$ and the risk of the action function $\phi$ at $\theta \in \Theta$ is the real number
\begin{equation}
\mbox{\tt Risk}_\infty(\theta;\phi) =
\E_{\mbox{\scriptsize \tt RDPG}} \left[
\mbox{\tt Loss} \left( \theta; \phi \left( 
E_\infty ( \ell_1,\ldots,\ell_n;\Delta_\infty,P_\infty) \right) \right) \right],
\label{eq:risk}
\end{equation}
where $\E_{\mbox{\scriptsize \tt RDPG}}$ denotes expectation with respect to the probability model $(A,L) \sim \mbox{\tt RDPG}(P_1,\ldots,P_g)$.

In practice we only observe $m_1,\ldots,m_N \iid P_\infty$, in which case we replace the original latent positions $\ell_1,\ldots,\ell_n$ with estimated latent positions $\hat{\ell}_1,\ldots,\hat{\ell}_n$, the Riemannian distance function $\Delta_\infty$ with a shortest path dissimilarity function $\Delta^+_N$, and the true sampling distribution $P_\infty$ with the empirical distribution $\hat{P}_N$.  The risk of $\phi$ at $\theta$ is then the random variable
\begin{equation}
\mbox{\tt Risk}_N(\theta;\phi) =
\E_{\mbox{\scriptsize \tt RDPG}} \left[
\mbox{\tt Loss} \left( \theta; \phi \left( 
E_N ( \hat{\ell}_1,\ldots,\hat{\ell}_n;\Delta^+_N,\hat{P}_N) \right) \right) \right].
\label{eq:riskN}
\end{equation}
In the next section we will demonstrate that, for each $\theta \in \Theta$, (\ref{eq:riskN}) converges in probability to (\ref{eq:risk}) as $N \rightarrow \infty$.

\section{Convergence}
\label{converge}

Our convergence analysis concatenates four distinct investigations: \\

{\bf 1}.  Convergence of the estimated latent positions to the true latent positions.  Here we rely on previous investigations of ASE in \cite{VL&etal:2017,2toInf:2019,LSM:2018},
which establish a strong form of convergence:
as $k \rightarrow \infty$, the probability increases to certainty that the maximal estimation error will be smaller than any specified threshold.
It is this property of ASE that compels its use.
We do not claim (or even believe) that ASE is the optimal way to estimate latent positions, but---at least for now---it is the only method we know to have the desired convergence property. \\

{\bf 2}.  Convergence of the computed dissimilarities between estimated latent positions to the Riemannian distances between the true latent positions.  Here we leverage results on graph approximations to geodesic curves, obtained in 
\cite{Bernstein&etal:2000} for studying the manifold learning procedure Isomap \cite{isomap:2000}.  Note that Step 4 of our decision rules corresponds to the first two steps of Isomap, which approximate Riemannian distance with shortest path distance.  Unlike Isomap, however, the data points do not lie on the manifold, but in a small tubular neighborhood of it. \\

{\bf 3}.  Convergence of the Euclidean representation to an interpretable limit.  Note that Step 5 of our decision rules corresponds to Step 3 of Isomap, except that, following a suggestion in \cite{mwt:isomap}, we replace classical multidimensional scaling with minimization of Kruskal's \cite{kruskal:1964a} raw stress criterion.  This substitution allows us to exploit the continuous multidimensional scaling framework explicated in \cite{mwt:ContinMDS}, as well as the convergence results therein. \\

{\bf 4}. Convergence of the risk function to an interpretable limit.  Here we deploy conventional continuity arguments.


\subsection{Of the estimated latent positions}

Because we are exclusively concerned with RDPGs, we need not consider generalized ASE.  Instead, we replace $\sigma_i^2 = |\lambda_i|$ in Definition \ref{def:ASE} with $\sigma_i^2 = \max(\lambda_i,0)$, resulting in traditional ASE.
We can then invoke the following consistency result, which
guarantees a form of {\em uniform}\/ convergence of the estimated latent positions to the true latent positions, a property that is crucial for our analysis.

The edge probabilities of an RDPG depend on its latent positions only through their pairwise inner products.  
If $X$ is the $k \times q$ latent position matrix and
$W$ is any $q \times q$ orthogonal matrix, then $(XW)(XW)^\top = XWW^\top X^\top = XX^\top$; hence, the latent positions $X$ and $XW$ have the same edge probabilities, i.e., the latent positions are not identifiable.
Even so, they can be consistently estimated in the following sense.
\begin{theorem}[\cite{VL&etal:2017,2toInf:2019,LSM:2018}]
Let
\[
X_k = \left[ \begin{array}{c|c|c} x_1 & \cdots & x_k \end{array} \right]^\top,
\]
with $\mbox{\rm rank}(X_k)=q$ for all sufficiently large $k$.
Set
$g_k = \max_i \sum_j x_{i}^\top x_{j}$,
the maximum expected degree of the adjacency matrix $A_k$, which increases more rapidly than $\log^2 k$ as $k \rightarrow \infty$.
Suppose that
$A_k|X_k \sim \mbox{\tt RDPG}(X_k)$ and
let $\hat{x}_{k,1},\ldots,\hat{x}_{k,v}$ denote the ASE of $A_k$ in $\Re^q$.
Then there exists $C>0$ and a sequence of $q \times q$ orthogonal matrices $W_k$ such that
\begin{equation}
\lim_{k \rightarrow \infty} P_{\mbox{\scriptsize \tt RDPG}} \left(
\max_{i=1,\ldots,k} \left\|
W_k \hat{x}_{k,i} - x_{i} 
\right\| \leq C \left( q/g_k \right)^{1/2} \log^2 k 
\right) = 1.
\label{eq:2inf}
\end{equation}
\label{thm:2inf}
\end{theorem}
We will set $k=n+N$ and apply Theorem~\ref{thm:2inf} to the augmented adjacency matrix $A_N^+$, letting $N \rightarrow \infty$ while $n$ remains fixed.

The orthogonal matrices $W_k$ that appear in (\ref{eq:2inf}) are unknown; however, 
Step 3 of our randomized decision rule replaces $\hat{\ell}_1,\ldots,\hat{\ell}_n,\hat{m}_1,\ldots,\hat{m}_N \in \Re^k$ 
with an $(n+N) \times (n+N)$ dissimilarity matrix $\Delta_N$ that is invariant under rotation.  In consequence, we can assume that each set of estimated latent positions has been suitably rotated, obviating the rotations in (\ref{eq:2inf}).  Thus, we can reinterpret Theorem \ref{thm:2inf} as 
\begin{cor}
Given any probability $\alpha>0$ and any error threshold $\sigma>0$, 
there exists $N(\alpha,\sigma)$ such that $N \geq N(\alpha,\sigma)$ entails
\begin{equation}
P_{\mbox{\scriptsize \tt RDPG}} \left( 
 \max_{i=1,\ldots,n} 
\left\| \hat{\ell}_i - \ell_i \right\| \leq \sigma,
\max_{i=1,\ldots,N}
\left\| \hat{m}_i - m_i \right\| \leq \sigma \right)
\geq 1-\alpha.
\label{eq:2inf.cor}
\end{equation}
\label{cor:2inf}
\end{cor}

\subsection{Of the dissimilarities}
\label{dissim}

As in \cite{Bernstein&etal:2000}, we construct localization graphs from the estimated latent positions.
\begin{defn}
Given a finite set $V \subset \Re^q$ and $\epsilon_2 > \epsilon_1 > 0$, let $\G(\epsilon_1,\epsilon_2)$ denote a weighted graph with vertex set $V$.  An edge between two vertices is present if the Euclidean distance between them is no greater than $\epsilon_1$ and absent if the Euclidean distance between them is no less than $\epsilon_2$.  The weights assigned to the edges are the pairwise Euclidean distances.  If $\G(\epsilon_1,\epsilon_2)$ is connected, then we say that $\G(\epsilon_1,\epsilon_2)$ localizes $V$.
\label{def:localG}
\end{defn}
We emphasize that the construction of $\G(\epsilon_1,\epsilon_2)$ is invariant under rotation.  Hence, we can construct $\G(\epsilon_1,\epsilon_2)$ from the observed $\hat{\ell}_1,\ldots,\hat{\ell}_n,\hat{m}_1,\ldots,\hat{m}_N \in \Re^q$, but we can study the properties of $\G(\epsilon_1,\epsilon_2)$ as though these vectors had been rotated to align with ${\ell}_1,\ldots,{\ell}_n,{m}_1,\ldots,{m}_N$, as in Corollary \ref{cor:2inf}.

Suppose that $\G=\G(\epsilon_1,\epsilon_2)$ localizes $V$ and
let $\Delta_\G$ denote shortest path distance on $\G$.  The following inequalities show that, if $V \subset \M_2 \subset \Re^q$, then $\Delta_\G$ ``is a good approximation to'' Riemannian distance on $\M_2$ ``under favorable circumstances'' \cite[p.\ 3]{Bernstein&etal:2000}
for all pairs of vertices.  See Appendix~A for discussion of the manifold parameters {\em minimum radius of curvature}\/ and {\em minimum branch separation}.

\begin{theorem}[\cite{Bernstein&etal:2000}, Main Theorem A]
Let $\M_2 \subset \Re^q$ be a smooth, compact, and connected Riemannian manifold with minimum radius of curvature $r_0(\M_2)$, minimum branch separation $s_0(\M_2)$, and Riemannian distance function $\Delta_{\M_2}$.
Given $\lambda_1, \lambda_2 \in (0,1)$, suppose that
\begin{itemize}

\item $\epsilon_1,\epsilon_2$ satisfy
$0 < \epsilon_1 < 
\epsilon_2 \leq \min \left\{ s_0 \left( \M_2 \right), \frac{4}{\pi}r_0 \left( \M_2 \right) \sqrt{6\lambda_1} \right\}$;

\item for $\delta_1$ such that $4 \delta_1 \leq \lambda_2 \epsilon_1$, the set $\vec{x} = \{ x_1,\ldots,x_k \} \subset \M_1 \subset \M_2$ satisfies the $\delta_1$-sampling condition in $\M_1$, i.e., for every $y \in \M_1$, there exists $x_i \in \vec{x}$ such that $\Delta_{\M_2}(y,x_i) \leq \delta_1$; and

\item $\G=\G(\epsilon_1,\epsilon_2)$ localizes $\vec{x}$.

\end{itemize}
Then 
$\left( 1-\lambda_1 \right) \Delta_{\M_2}(x_i,x_j) \leq
\Delta_\G(x_i,x_j) \leq
\left( 1+\lambda_2 \right) \Delta_{\M_2}(x_i,x_j)$
for all $x_i,x_j \in \vec{x}$.
\label{thm:SPD}
\end{theorem}

\subparagraph{Remark}  Main Theorem A in \cite{Bernstein&etal:2000} takes $\M_1 = \M_2$.  Our version is obtained by weakening the statement of the $\delta$-sampling condition in their Theorem~2.

\bigskip

Notice that there is an easy way to extend $\Delta_\G$ from a distance function on $\vec{x} \times\vec{x}$ to a dissimilarity function on $\M_1 \times \M_1$.  Given any $(y_1,y_2) \in \M_1 \times \M_1$, define ${\Delta}^+_\G(y_1,y_2)$ to be the shortest path distance on the graph $\G^+$ with vertex set $\{ x_1,\ldots,x_k,y_1,y_2 \}$.
If the conditions in Theorem~\ref{thm:SPD} obtain for $\G$, then they also obtain for $\G^+$ and we conclude that the inequalities
\begin{equation}
\left( 1-\lambda_1 \right) \Delta_\infty \left( y_1,y_2 \right) 
\leq {\Delta}^+_\G \left( y_1,y_2 \right) \leq
\left( 1+\lambda_2 \right) \Delta_\infty \left( y_1,y_2 \right)
\label{eq:SPD}
\end{equation}
hold for all $y_1,y_2 \in \M_1$.

For our purposes, the limitation of Theorem~\ref{thm:SPD} is that it only considers data sets that lie {\em on}\/ the manifold of interest, $x_1,\ldots,x_k \in \M_2$.  Our data sets comprise estimated latent positions, which are only guaranteed to lie {\em near}\/ $\M$.  But suppose that these estimates, $\hat{x}_1,\ldots,\hat{x}_k$, lie within a small tubular radius $\sigma_1$ of $\M$, i.e., on the tubular manifolds $\M_1 = \M_{\sigma_1} \subset \M_{\sigma_2} = \M_2$.  (See Appendix~A for discussion of tubular radii and tubular manifolds.)  Then we can apply Theorem~\ref{thm:SPD} and its extension (\ref{eq:SPD}) to the tubular manifolds, obtaining the following result.

\begin{cor}
Let $\M \subset \Re^q$ be a smooth, compact, and connected Riemannian manifold with minimum radius of curvature $r_0$, minimum branch separation $s_0$, and Riemannian distance $\Delta_\infty = d_0$.  Let $\sigma_1 \leq \sigma_2$ be tubular radii of $\M$ and let $d_\sigma$ denote Riemannian distance on $\M_\sigma$.
Given $\lambda_1, \lambda_2 \in (0,1)$, 
Suppose that
\begin{itemize}

\item $\epsilon_1,\epsilon_2$ satisfy
$0 < \epsilon_1 < 
\epsilon_2 \leq \min \left\{ s_0-2 \sigma_2, \frac{4\sigma_2}{\pi} \sqrt{6\lambda_1} \right\}$;

\item $x_1,\ldots,x_k \in \M$ satisfy the $\delta$-sampling condition in $\M$ with 
$4 (\delta+\sigma_1) \leq \lambda_2 \epsilon_1$; 

\item for each $x_i$, there exists $\hat{x}_i \in \Re^q$ such that $\| \hat{x}_i-x_i \| \leq \sigma_1$; and

\item $\G=\G(\epsilon_1,\epsilon_2)$ localizes $\hat{x}_1,\ldots,\hat{x}_k$.

\end{itemize}
Then the inequalities
\begin{equation}
\left( 1-\lambda_1 \right) d_{\sigma_2} \left( y_1,y_2 \right) 
\leq \Delta^+_\G \left( y_1,y_2 \right) \leq
\left( 1+\lambda_2 \right) d_{\sigma_2} \left( y_1,y_2 \right)
\label{eq:SPDsigma}
\end{equation}
hold for all $y_1,y_2 \in \M_{\sigma_1}$, hence for all $y_1,y_2 \in \M$.
\label{cor:SPD}
\end{cor}

\subparagraph{Proof}
Because $\sigma$ is a tubular radius of $\M$ and each $\hat{x}_i \in \Re^q$ lies within $\sigma$ of $x_i \in \M$,
the observed data set $\hat{x}_1,\ldots,\hat{x}_k$ lies in the tubular manifold $\M_1=\M_{\sigma_1}$.  From the Appendix, the minimum radius of curvature of the tubular manifold $\M_2 = \M_{\sigma_2}$ is not less than $\sigma_2$ and the minimum branch separation of $\M_2$ is $s_0-2\sigma_2$.  Substituting these quantities into the assumed upper bound on $\epsilon_2$ in Theorem~\ref{thm:SPD} results in 
\[
\epsilon_2 \leq \min \left\{ s_0-2 \sigma_2, \frac{4\sigma_2}{\pi} \sqrt{6\lambda_1} \right\}.
\]

It remains to show that $\hat{x}_1,\ldots,\hat{x}_k$ satisfy a suitable sampling condition in $\M_1$.  
For any $\hat{y} \in \M_1$, there exists $y \in \M$ such that $d_{\sigma_2} (\hat{y},y) = \| \hat{y}-y \| \leq \sigma_1$.  By hypothesis, there exists an $x_i$ such that $d_{\sigma_2} (y,x_i) \leq d_0 (y,x_i) \leq \delta$; hence, by the triangle inequality,
$d_{\sigma_2} (\hat{y},x_i) \leq \sigma_1+\delta$ and $\hat{x}_1,\ldots,\hat{x}_k$ satisfy the $(\delta+\sigma_1)$-sampling condition in $\M_1$.  Replacing the assumption $4 \delta_1 \leq \lambda_2 \epsilon_1$ in Theorem~\ref{thm:SPD} with the assumption $4 (\delta+\sigma_1) \leq \lambda_2 \epsilon_1$ completes the application of Theorem~\ref{thm:SPD}. \hfill $\Box$

\medskip

To approximate the Riemannian distance function $\Delta_\infty : \M \times \M \rightarrow \Re$ with the dissimilarity function $\Delta_\G : \M \times \M \rightarrow \Re$, we require inequalities analogous to (\ref{eq:SPDsigma}) with $d_0=\Delta_\infty$ replacing $d_\sigma$.
\begin{theorem}
Let $\M \subset \Re^q$ be a smooth, compact, and connected Riemannian manifold with minimum radius of curvature $r_0$, minimum branch separation $s_0$, and Riemannian distance function $\Delta_\infty = d_0$.  Given $\rho \in (1,2)$, let $\lambda_1 = 1-1/\sqrt{\rho}$ and 
$b = \min \{ s_0, (4/\pi) r_0 \sqrt{6 \lambda_1} \}$. Let
$0 < \sigma_1 \leq \sigma_2 \leq \lambda_1 b r_0 / \mbox{\tt diam}(\M)$ be tubular radii of $\M$ and suppose that 
\begin{itemize}

\item $\epsilon_1,\epsilon_2$ satisfy
$0 < \epsilon_1 < 
\epsilon_2 \leq \min \left\{ s_0-2 \sigma_2, \frac{4\sigma_2}{\pi} \sqrt{6\lambda_1} \right\}$;

\item $x_1,\ldots,x_k \in \M$ satisfy the $\delta$-sampling condition in $\M$ with
$4 (\delta+\sigma_1) \leq (\rho-1) \epsilon_1$;

\item for each $x_i$, there exists $\hat{x}_i \in \Re^q$ such that
$\| \hat{x}_i-x_i \| \leq \sigma_1$; and

\item $\G=\G(\epsilon_1,\epsilon_2)$ localizes $\hat{x}_1,\ldots,\hat{x}_k$. 

\end{itemize}
Then the inequalities
\begin{equation}
\frac{1}{\rho} \, \Delta_\infty \left( y_1,y_2 \right) 
\leq \Delta^+_\G \left( y_1,y_2 \right) \leq
\rho \, \Delta_\infty \left( y_1,y_2 \right)
\label{eq:SPD0}
\end{equation}
hold for all $y_1,y_2 \in \M$.
\label{thm:SPD0}
\end{theorem}

\subparagraph{Proof}
First, using $\lambda_1$ and $\lambda_2 = \rho-1$ we apply Corollary~\ref{cor:SPD} to obtain
\[ 
\frac{1}{\sqrt{\rho}} \,
d_{\sigma_2} \left( y_1,y_2 \right) 
\leq \Delta^+_\G \left( y_1,y_2 \right) \leq
\rho \, d_{\sigma_2} \left( y_1,y_2 \right).
\]
As $d_{\sigma_2} ( y_1,y_2 ) \leq d_0 ( y_1,y_2 )$, we immediately obtain $\Delta^+_\G ( y_1,y_2 ) \leq
\rho \, d_0 ( y_1,y_2 )$.

Next, partition $\M \times \M$ into
$B = \{ ( y_1,y_2 ) : \| y_1-y_2 \| < b \}$
and $B^c$.    
For $(y_1,y_2) \in B$, apply Lemma~\ref{lm:euclid} with $\lambda = \lambda_1$ to obtain
$\left( 1-\lambda_1 \right) d_0 \left( y_1,y_2 \right) \leq 
\left\| y_1 - y_2 \right\|$,
hence
\[
\frac{1}{\rho} \, d_0 \left( y_1,y_2 \right) = 
\frac{1-\lambda_1}{\sqrt{\rho}} \,  d_0 \left( y_1,y_2 \right) \leq 
\frac{1}{\sqrt{\rho}} \, \left\| y_1 - y_2 \right\|  \leq 
\frac{1}{\sqrt{\rho}} \, d_{\sigma_2} \left( y_1,y_2 \right) \leq 
\Delta^+_\G \left( y_1,y_2 \right).
\]
For $(y_1,y_2) \in B^c$, 
\[
d_0 \left( y_1,y_2 \right) \geq
d_{\sigma_2} \left( y_1,y_2 \right) \geq
\left\| y_1 - y_2 \right\| \geq b > 0.
\]
Applying Lemma~\ref{lm:tubular} with $\sigma = \sigma_2$, we obtain
\[
d_0 \left( y_1,y_2 \right) - 
d_{\sigma_2} \left( y_1,y_2 \right) \leq \sigma_2 \, \mbox{\tt diam}(\M)/r_0,
\]
hence
\begin{eqnarray*}
\frac{d_{\sigma_2} \left( y_1,y_2 \right)}{d_0 \left( y_1,y_2 \right)}
 & \geq & 
1 - \frac{\sigma_2 \, \mbox{\tt diam}(\M)}{d_0 \left( y_1,y_2 \right) r_0}
\geq 1 - \sigma_2 \, \frac{\mbox{\tt diam}(\M)}{b r_0} \\
 & \geq & 
1 - \frac{\lambda_1 b r_0}{\mbox{\tt diam}(\M)} \, \frac{\mbox{\tt diam}(\M)}{b r_0} = 1-\lambda_1 = 1/\sqrt{\rho},
\end{eqnarray*}
and finally
\[
\frac{1}{\rho} \, d_0 \left( y_1,y_2 \right) \leq
\frac{1}{\sqrt{\rho}} \, d_{\sigma_2} \left( y_1,y_2 \right) \leq
\Delta^+_\G \left( y_1,y_2 \right).
\]
\hfill $\Box$

\bigskip

Next we consider how to obtain $x_1,\ldots,x_k \in \M$ that satisfy the $\delta$-sampling condition in $\M$.
The following lemma is analogous to the Sampling Lemma in \cite{Bernstein&etal:2000}.
\begin{lemma}
Let $\M$ be a compact Riemannian manifold and let $P_\infty$ be a probability measure on $(\M,\B)$ with full support.
Suppose that $x_1,\ldots,x_k \iid P_\infty$ and let
$E_k$ denote the event that $x_1,\ldots,x_k$ satisfy the $\delta$-sampling condition, i.e., every $y \in \M$ lies within Riemannian distance $\delta$ of some $x_j$.  Then
$\lim_{k \rightarrow \infty} P_\infty(E_k)=1$.
\label{lm:sampling}
\end{lemma}

\subparagraph{Proof}
Let $\Delta_\infty$ denote the Riemannian distance function on $\M$.
For $r < \delta/2$, let $B_1,\ldots,B_{I(r)}$ be a finite covering of $\M$ by geodesic balls, i.e., each 
\[
B_i = \left\{ y \in \M : \Delta_\infty \left( y,y_i \right) < r \right\}
\]
for some $y_i \in \M$.
If each $B_i$ contains at least one $x_j$, then $E_k$ obtains.

Let $p_{\min}$ denote the strictly positive minimum of the $P_\infty(B_i)$.  Then
\begin{eqnarray*}
\lefteqn{P_\infty \left( \mbox{every $B_i$ contains an $x_j$} \right)} \\
 & = & 1- P_\infty \left( \mbox{some $B_i$ contains no $x_j$} \right)  
   \geq  1- \sum_{i=1}^{I(r)} P_\infty \left( \mbox{$B_i$ contains no $x_j$} \right) \\
  & = & 1- \sum_{i=1}^{I(r)} \prod_{j=1}^k P_\infty \left( x_j \not\in B_i \right) 
   =  1- \sum_{i=1}^{I(r)} \prod_{j=1}^k \left(
  1 - P_\infty \left( B_i \right) \right) \\
  & \geq & 1- \sum_{i=1}^{I(r)} \prod_{j=1}^k \left(
  1 - p_{\min} \right) 
   =  1-I(r) \left(
  1 - p_{\min} \right)^k,
\end{eqnarray*}
which tends to $1$ as $k \rightarrow \infty$.
\hfill $\Box$

\bigskip

We have now collected the results needed to establish convergence of the computed dissimilarities between estimated latent positions to the Riemannian distances between the true latent positions.  The mode of convergence is quite strong and requires careful explication.

\begin{defn}
Let $\Delta_\infty$ and $\Delta_{1},\Delta_{2},\ldots$ denote dissimilarity functions on $\M \times \M$ that are strictly positive if $y_1 \neq y_2$.  Let
\[
\mu \left( \Delta_{j},\Delta_\infty \right) = \log \, \inf
\left\{ r : \frac{1}{r} \leq \frac{\Delta_{j} \left( y_1,y_2 \right)}{\Delta_\infty \left( y_1,y_2 \right)} \leq r \mbox{ for all } y_1 \neq y_2 \right\}.
\]
If
$\lim_{j \rightarrow \infty} \mu \left( \Delta_{j},\Delta_\infty \right) = 0$,
then we say that $\Delta_{j}$ converges in ratio to $\Delta_\infty$.
\label{def:ratio}
\end{defn}

The metric $\mu$ is motivated in \cite[Section~5]{mwt:ContinMDS}.
It is essentially the supremum norm of the difference in logarithmic dissimilarity.  As $\M$ is compact, the dissimilarity functions in question are bounded and convergence in ratio implies uniform convergence.

\begin{theorem}
Suppose that the latent positions of an RDPG lie on a smooth connected compact Rienmannian manifold $\M \subset \Re^q$ with minimum radius of curvature $r_0$, minimum branch separation $s_0$, and Riemannian distance function $\Delta_\infty$.  Let $P_\infty$ be a probability measure on $(\M,\B)$ with full support and suppose that $m_1,m_2,\ldots \iid P_\infty$. 

For $j=1,2,\ldots$, consider the RDPG with latent positions $\ell_1,\ldots,\ell_n,m_1,\ldots,m_{N(j)}$ and ASE-estimated latent positions $\hat{\ell}_1,\ldots,\hat{\ell}_n,\hat{m}_1,\ldots,\hat{m}_{N(j)}$.  Let $\G_{N(j)}$ denote any $\G(\epsilon_1,\epsilon_2)$ that localizes $\hat{\ell}_1,\ldots,\hat{\ell}_n,\hat{m}_1,\ldots,\hat{m}_{N(j)}$.  Let $\Delta^+_{N(j)}$ denote shortest path distance on $\G_{N(j)}$, extended to a dissimilarity function on $\M$.  Then there exist sequences $\{ N(j) \}$, $\{ \epsilon_1(j) \}$, and $\{ \epsilon_2(j) \}$ such that $\Delta^+_{N(j)}$ almost surely converges in ratio to $\Delta_\infty$.
\label{thm:SPDconverge}
\end{theorem}

\subparagraph{Proof}
Let $\{ \alpha_j \}$ be a decreasing sequence of strictly positive probabilities such that $\{ \alpha_j \}$ is summable, i.e., $\sum_{j=1}^\infty \alpha_j < \infty$.  Let $\{ \rho_j \} \subset (1,2)$ be a strictly decreasing sequence of ratios that converge to unity.  For example, we might set $\alpha_j = 1/(j+1)^2$ and $\rho_j = 1+1/(j+1)$.
For each $j$, let
\begin{eqnarray*}
\lambda_1(j) = 1-1/\sqrt{\rho(j)} & \mbox{and} &
b(j) = \min \left\{ s_0, \frac{4r_0}{\pi} \sqrt{6 \lambda_1(j)} \right\}.
\end{eqnarray*}
Let $0 < \sigma_2(j) \leq \lambda_1(j)br_0/\mbox{\tt diam}(\M)$
be a tubular radius of $\M$ and choose
$\epsilon_2(j)$ in
\[
\left( 0,
\min \left\{
s_0-2\sigma_2(j), \frac{4\sigma_2(j)}{\pi} \sqrt{6 \lambda_1(j)} \right\} \right].
\]
Choose
$0 < \sigma_1(j) < \sigma_2(j)$ and
$0 < \epsilon_1(j) < \epsilon_2(j)$ such that $\sigma_1(j)/\epsilon_1(j) \leq (\rho(j)-1)/8$.  Then the hypotheses of Theorem~\ref{thm:SPD0}  are satisfied if
\begin{enumerate}

\item each $\hat{\ell}_i$ lies with $\sigma_1(j)$ of $\ell_j$ and each $\hat{m}_i$ lies within $\sigma_1(j)$ of $m_i$; and
 
\item $\ell_1,\ldots,\ell_n,m_1,\ldots,m_{N(j)}$ satisfy the $\delta(j)$-sampling condition for $\delta(j) < \sigma_1(j)$.

\end{enumerate}
Corollary~\ref{cor:2inf} allows us to choose $N(j)$ sufficiently large that the first requirement obtains with probability at least $1-\alpha_j$.  Setting $x_i=m_i$ and $k=N$, Lemma~\ref{lm:sampling} allows us to choose $N(j)$ sufficiently large that $m_1,\ldots,m_{N(j)}$, hence $\ell_1,\ldots,\ell_n,m_1,\ldots,m_{N(j)}$, satisfy the $\delta(j)$-sampling condition with probability at least $1-\alpha_j$.  Choosing the larger of these $N(j)$, the hypotheses of Theorem~\ref{thm:SPD0} are satisfied  with probability at least $1-2\alpha_j$.  As the sequence of error probabilities $\{ 2 \alpha_j \}$ is summable, the probability that the hypotheses of Theorem~\ref{thm:SPD0} fail to hold infinitely often is zero by the Borel-Cantelli Lemma.  Hence, for sufficiently large $j$, we have
\[
\frac{1}{\rho_j} \, \Delta_\infty \left( y_1,y_2 \right) 
\leq \Delta^+_{N(j)} \left( y_1,y_2 \right) \leq
\rho_j \, \Delta_\infty \left( y_1,y_2 \right)
\]
for all $y_1,y_2 \in \M$.
As $\rho_j \downarrow 1$, $\Delta^+_{N(j)}$ almost surely converges in ratio to $\Delta_\infty$.
\hfill $\Box$

\subsection{Of the Euclidean representations}

Information about the manifold structure of $\M$ is encoded in the  dissimilarity functions $\Delta^+_{N(j)}$.
The randomized decision rules that concern us construct Euclidean representations of this information.  We now argue 
that these representations converge to meaningful limits.

A popular way of embedding a $k \times k$ dissimilarity matrix $\Delta = [ \delta_{rs} ]$ in $\Re^d$ involves minimizing Kruskal's raw stress criterion \cite{kruskal:1964a},
\[
\min \sigma_k(Z;\Delta) = \sum_{r,s=1}^k
w_{rs} \left[ \left\| z_r-z_s \right\| - \delta_{rs}
\right]^2
\]
to obtain a configuration matrix $Z = [ z_1 | \cdots z_k ]^\top$.
Notice that $\sigma_k$ is invariant under isometric transformations of $Z$, so one can also work directly with $D(Z) = [ \left\| z_r-z_s \right\|$, the matrix of pairwise interpoint Euclidean distances in $\Re^d$.  If one knows such a $D$, then one can always recover a $Z$ for which $D(Z)=D$.  Using the raw stress criterion allows us to exploit the convergence analyses in \cite{mwt:ContinMDS}.

\subsubsection{Partial Graph Embedding}

The analysis of PGE is relatively straightforward, as the same $n$ latent positions are embedded for each $j$ and only the $n \times n$ dissimilarity matrices associated with those positions vary.
For each $N(j)$, multidimensional scaling (MDS) constructs a Euclidean representation of the $n \times n$ dissimilarity matrix $\mbox{\tt PGE}_{N(j)}$, obtained by evaluating $\Delta^+_{N(j)}$ at each pair of $\hat{\ell}_1,\ldots,\hat{\ell}_n$ and resulting in a configuration of $n$ points in $\Re^d$.

Set $w_{rs}=1/n^2$, the case of equal weights.
Following \cite[Section~3]{mwt:ContinMDS},
an $n \times n$ matrix $D=[d_{ij}]$ is EDM-1 if and only if there exists a dimension $p$ and points $z_1,\ldots,z_n \in \Re^p$ such that $d_{ij} = \| z_i-z_j \|$.  The smallest such $p$ is the embedding dimension of $D$.
If $D$ denotes an $n \times n$ EDM-1 matrix of embedding dimension $\leq d$, then we can rewrite $\sigma_n$ as
\[
\sigma_n(\Delta,D) = \frac{1}{n^2} \| D-\Delta \|_F^2,
\]
where $\| \cdot \|_F$ is the Frobenius norm.  The problem of minimizing the raw stress criterion can thus be reformulated as the problem of minimizing $\sigma_n$ as $D$ varies in the closed cone of all $n \times n$ EDM-1 matrices of embedding dimension $\leq d$.
Let ${\tt Min}(\Delta)$ denote the set of such $D$ that are global minimizers of $\sigma_n(\cdot;\Delta)$.

Let $\mbox{\tt PGE}_\infty = [ \Delta_\infty(\ell_r,\ell_s) ]$ denote the $n \times n$ dissimilarity matrix of pairwise Riemannian distances between the $\ell_1,\ldots,\ell_n \in \M$. 
Under the assumptions of Theorem~\ref{thm:SPDconverge},
\[
\lim_{j \rightarrow \infty} 
\left\|  \mbox{\tt PGE}_{N(j)}
- \mbox{\tt PGE}_\infty \right\|_F = 0;
\]
hence, we can apply \cite[Theorem~2]{mwt:ContinMDS} to conclude that 
\begin{theorem}
With probability one,
\begin{itemize}

\item[(a)] any sequence of $n \times n$ EDM-1 matrices $D_{N(j)} \in {\tt Min}(\mbox{\tt PGE}_{N(j)})$ has an accumulation point in the topology of the Frobenius norm, and 

\item[(b)] if the $n \times n$ EDM-1 matrix $D_\infty$ is an accumulation point of $\{ D_{N(j)} \in {\tt Min}(\mbox{\tt PGE}_{N(j)}) \}$ in the topology of the Frobenius norm, then $D_\infty \in {\tt Min}(\mbox{\tt PGE}_\infty)$.

\end{itemize}
\label{thm:PGE}
\end{theorem}

\subsubsection{Complete Graph Embedding}

For each $N(j)$, MDS constructs a Euclidean representation of the $[n+N(j)] \times [n+N(j)]$ dissimilarity matrix $\mbox{\tt CGE}_{N(j)}$, obtained by evaluating $\Delta^+_{N(j)}$ at each pair of ASE-estimated latent positions and resulting in a configuration of $n+N(j)$ points in $\Re^d$.  But in what sense does a sequence of configurations converge if the numbers of points in the configurations tend to infinity?  

The continuous MDS framework proposed in \cite{mwt:ContinMDS} replaces the configuration matrix with an embedding function and
replaces the dissimilarity matrix with a (dissimilarity function, probability measure) pair.  Let ${\tt mds} : {\mathcal M} \rightarrow \Re^d$ denote a Borel-measurable embedding function,  let $\Delta : \M \times \M \rightarrow \Re$ denote a Borel-measurable dissimilarity function, and let $P$ denote a probability measure on $(\M,\P)$.
Then the continuous raw stress criterion is defined by
\[
\sigma \left( {\tt mds}; (\Delta, P)  \right) =
\int_{\mathcal M} \int_{\mathcal M}
\left[ \left\| {\tt mds} \left( y_1 \right) -
{\tt mds} \left( y_2 \right) \right\| 
- \Delta \left( y_1,y_2 \right) \right]^2
P \left( dy_1 \right) P \left( dy_2 \right) .
\]
Notice that, if $P$ is a discrete probability measure supported on $k$ points, then one recovers the traditional weighted raw stress criterion.

Whatever the support of $P$, one can interpret an ${\tt mds}$ that minimizes $\sigma$ as an embedding function that constructs an optimal $d$-dimensional Euclidean representation of $(\Delta,P)$.  We do not claim that such embedding functions can actually be computed when $P$ has full support.  Even if $\M$ were known, the resulting problem in the calculus of variations would likely be intractable.  What we claim is that the intent of a continuous MDS problem is easily understood and interpreted.

Observe that any function $D : {\mathcal M} \times {\mathcal M} \rightarrow \Re$ defined by
\[
D \left( y_1,y_2 \right) = \left\| {\tt mds} \left( y_1 \right) -
{\tt mds} \left( y_2 \right) \right\|
\]
is a pseudometric on ${\mathcal M}$.  Following \cite[Section~4]{mwt:ContinMDS}, we refer to pseudometrics of this form as $d$-dimensional Euclidean pseudometrics.  Let ${\tt Min}(\Delta,P)$ denote the set of $d$-dimensional Euclidean pseudometrics that are global minimizers of $\sigma(\cdot;\Delta,P)$.

Assume that $\M$ is a compact metric space, that $\Delta_\infty$ metrizes $\M$, and that $m_1,m_2,\ldots \iid P_\infty$.  Let $\hat{P}_{N(j)}$ denote the empirical distribution of $m_1,\ldots,m_{N(j)}$, which converge weakly to $P_\infty$.  Notice that incorporating $\ell_1,\ldots,\ell_n$ into the $\hat{P}_{N(j)}$ does not affect convergence to $P_\infty$.  Because $\M$ is compact, convergence in ratio of $\Delta^+_{N(j)}$ to $\Delta_\infty$ implies that the $\Delta^+_{N(j)}$ are uniformly bounded and converge pointwise.
We can therefore apply \cite[Theorem~3]{mwt:ContinMDS} to conclude that 
\begin{theorem}
With probability one,
\begin{itemize}

\item[(a)] any sequence of $D_{N(j)} \in {\tt Min}(\Delta^+_{N(j)},\hat{P}_{N(j)})$ has an accumulation point in the topology of $L^2(P_\infty)$ convergence, and 

\item[(b)] if $D_\infty$ is an accumulation point of $\{ D_{N(j)} \in {\tt Min}(\Delta^+_{N(j)},\hat{P}_{N(j)}) \}$ in the topology of $L^2(P_\infty)$ convergence, then $D_\infty \in {\tt Min}(\Delta_\infty,P_\infty)$.

\end{itemize}
\label{thm:CGE}
\end{theorem}

\subsubsection{CGE with ALE}

The convergence in ratio established in Section~\ref{dissim} is much stronger than the pointwise convergence required by Theorem~\ref{thm:ALE}.  In fact, \cite[Section~5]{mwt:ContinMDS} uses convergence in ratio of the dissimilarity functions to establish uniform convergence of the Euclidean pseudometrics for a modified embedding technique, {\em Approximate Lipschitz Embedding}\/ (ALE).  ALE minimizes $\sigma(\cdot; \Delta,P)$ subject to the {\em approximate Lipschitz constraints}
\[
P \left( D \left( y_1,y_2 \right) =
\left\| \mbox{\tt mds}\left( y_1 \right) -
\mbox{\tt mds}\left( y_2 \right) \right\| \leq
K \Delta \left( y_1,y_2 \right) \right) = 1.
\]
A plausible algorithm for ALE when $P$ concentrates on $y_1,\ldots,y_k$ is described in \cite[Appendix]{mwt:ContinMDS}.

Let ${\tt Min}_{\scriptsize \tt ALE}(\Delta,P)$ denote the set of $d$-dimensional Euclidean pseudometrics that correspond to global minimizers of ALE with parameters $\Delta$ and $P$.
We can then apply \cite[Theorem~5]{mwt:ContinMDS} to conclude that
\begin{theorem}
For any sequence of $D_{N(j)} \in {\tt Min}_{\scriptsize \tt ALE}(\Delta^+_{N(j)},\hat{P}_{N(j)})$,
there exists a corresponding sequence $\{ \bar{D}_{N(j)} \}$ such that
\begin{itemize}

\item[(a)]  if $y_1,y_2 \in \M$ lie in the support of $\hat{P}_{N(j)}$, then $\bar{D}_{N(j)}(y_1,y_2) = D_{N(j)}(y_1,y_2)$; and

\item[(b)] each $\bar{D}_{N(j)} \in {\tt Min}_{\scriptsize \tt ALE}(\Delta^+_{N(j)},\hat{P}_{N(j)})$.

\end{itemize}
With probability one,
\begin{itemize}

\item[(c)] the sequence $\{ \bar{D}_{N(j)} \}$ has an accumulation point in the topology of uniform convergence, and 

\item[(d)] if $\bar{D}_\infty$ is an accumulation point of $\{ \bar{D}_{(N(j)} \}$ in the topology of uniform convergence, then $\bar{D}_\infty \in {\tt Min}_{\scriptsize \tt ALE}(\Delta_\infty,P_\infty)$.
\end{itemize}
\label{thm:ALE}
\end{theorem}

\subsection{Of the risk functions}

For fixed $\theta \in \Theta$, we now consider the behavior of the random variables in (\ref{eq:riskN}) as $N \rightarrow \infty$,
relying on the following fact about convergence in distribution.
\begin{lemma}[\cite{billingsley:1971}, Corollary~3]
Let $S$ and $S^\prime$ be separable and complete metric spaces, each associated with its respective $\sigma$-field of Borel sets.
Suppose that $f : S \rightarrow S^\prime$ is measurable and that
$D_f$, the set of discontinuities of $f$, is a Borel subset of $S$.
Let $R_1,R_2,\ldots$ and $R_\infty$ be random elements with values in $S$ and let $Q_\infty$ denote the probability distribution of $R_\infty$.
If $R_k$ converges in distribution to $R_\infty$ as $k \rightarrow \infty$ and $Q_\infty(D_f)=0$, then $f(R_k)$ converges in distribution to $f(R_\infty)$.
\label{lm:WeakConv}
\end{lemma}
Three applications of Lemma~\ref{lm:WeakConv} produce the following result.
\begin{theorem}
Assume that $\ell_1,\ldots,\ell_n$ are random elements in $\M$ and that $\mbox{\tt Loss}(\theta; \cdot)$ is bounded.
Let $S$ denote the closed cone of $n \times n$ EDM-1 matrices with embedding dimension $\leq d$.
For each $\theta \in \Theta$, define $f_\theta : S \rightarrow \Re$ by $f_\theta(D) = \mbox{\tt Loss}(\theta;\phi(D))$.
Assume that each $f_\theta$ is measurable.

Consider any sequence of the decision rules described in Section~\ref{infer} that satisfies one of the following:
\begin{enumerate}

\item  Partial Graph Embedding.  $D_{N(j)} \in \mbox{\tt Min}( \mbox{\tt PGE}_{N(j)} )$ converges to
$D_\infty \in \mbox{\tt Min}( \mbox{\tt PGE}_\infty )$ as in Theorem~\ref{thm:PGE}, and the distribution of $E_\infty = D_\infty \in S$ assigns probability zero to the discontinuities of $f_\theta$.

\item  Complete Graph Embedding.  $D_{N(j)} \in \mbox{\tt Min}( \mbox{\tt CGE}_{N(j)} )$ converges to
$D_\infty \in \mbox{\tt Min}( \mbox{\tt CGE}_\infty )$ as in Theorem~\ref{thm:CGE}, and the distribution of $E_\infty = [ D_\infty ( \bar{\ell}_r,\bar{\ell}_s) ] \in S$ assigns probability zero to the discontinuities of $f_\theta$.

\item  CGE with ALE.  $\bar{D}_{N(j)} \in \mbox{\tt Min}( \mbox{\tt CGE}_{N(j)} )$ converges to
$\bar{D}_\infty \in \mbox{\tt Min}( \mbox{\tt CGE}_\infty )$ as in Theorem~\ref{thm:ALE}, and the distribution of $E_\infty = [ \bar{D}_\infty ( \bar{\ell}_r,\bar{\ell}_s) ] \in S$ assigns probability zero to the discontinuities of $f_\theta$.

\end{enumerate}
Then
\[
\mbox{\tt Risk}_{N(j)}(\theta;\phi) =
\E_{\mbox{\scriptsize \tt RDPG}} \left[
\mbox{\tt Loss} \left( \theta; \phi \left( 
E_{N(j)} ( \hat{\ell}_1,\ldots,\hat{\ell}_n; 
\Delta^+_{N(j)},\hat{P}_{N(j)}) \right) \right) \right]
\]
converges in probability to (\ref{eq:risk})
as ${j \rightarrow \infty}$.  (Note that the proper definitions of $E_{N(j)}$ above and $E_\infty$ in (\ref{eq:risk}) depend on which embedding strategy is used.)
\label{thm:risk}
\end{theorem}

\subparagraph{Proof}
In the case of partial graph embedding, Theorem~\ref{thm:PGE} asserts that the random matrices $E_{N(j)}$ converge almost surely, hence in distribution, to the random matrix $E_\infty$.  Applying Lemma~\ref{lm:WeakConv}, the random variables
\[
f_\theta(D_{N(j)}) =
\mbox{\tt Loss} \left( \theta; \phi \left( 
E_N ( \hat{\ell}_1,\ldots,\hat{\ell}_n; 
\Delta^+_{N(j)},\hat{P}_{N(j)}) \right) \right)
\]
converge in distribution to the random variable 
\[
f_\theta(D_\infty) =
\mbox{\tt Loss} \left( \theta; \phi \left( 
E_\infty ( \ell_1,\ldots,\ell_n;\Delta_\infty,P_\infty) \right) \right).
\]
As $\mbox{\tt Loss}(\theta,\cdot)$ is bounded, it follows that the random variables
$\E_{\mbox{\tt \scriptsize RDPG}} f_\theta(D_{N(j)})$
converge in distribution to the random variable $\E_{\mbox{\tt \scriptsize RDPG}} f_\theta(D_\infty)$, which is (\ref{eq:risk}).  As the limit random variable is constant, the convergence is also in probability.

In the case of complete graph embedding, Theorem~\ref{thm:CGE} asserts that the random functions $D_{N(j)}$ converge in $L^2(P_\infty)$ with probability one.  $L^2$ convergence does not imply pointwise convergence, but the proof of \cite[Theorem~3]{mwt:ContinMDS} derives $L^p$ convergence from the pointwise convergence of uniformly bounded functions.  It follows from the pointwise convergence of $D_{N(j)}$ to $D_\infty$ that the random matrices $E_{N(j)} = D_{N(j)} [ \bar{\ell}_r,\bar{\ell}_s ]$ converge almost surely, hence in distribution, to the random matrix $E_\infty = D_\infty [ \bar{\ell}_r,\bar{\ell}_s ]$.  The rest of the argument is identical to the case of PGE.

In the case of CGE with ALE, Theorem~\ref{thm:ALE} asserts that the random functions $\bar{D}_{N(j)}$ converge uniformly with probability one.  Uniform convergence does imply pointwise convergence, but such a strong result is not needed as the proof of $L^2$ convergence in \cite[Theorem~4]{mwt:ContinMDS} establishes pointwise convergence. 
Either way, it follows that the random matrices $E_{N(j)} = \bar{D}_{N(j)} [ \bar{\ell}_r,\bar{\ell}_s ]$ converge almost surely, hence in distribution, to the random matrix $E_\infty = \bar{D}_\infty [ \bar{\ell}_r,\bar{\ell}_s ]$.  Again, the rest of the argument is identical to the case of PGE.
\hfill $\Box$

\subparagraph{Remark}  The condition that the distribution of $E_\infty$ assigns probability zero to the discontinuities of $f_\theta$ is difficult to check.  In the future, we hope to find more practical conditions that imply it.

\section{Discussion}
\label{discuss}

Random dot product graphs are popular generative probability models in network machine learning.  Furthermore, RDPGs with latent positions on a low-dimensional Riemannian support manifold provide a useful idealization of the more general class of latent position models.  To perform statistical inference in this setting, it is natural to inquire how one might achieve (some of) the potential benefits of restricted inference.

This report extends the studies initiated in 
\cite{LSM:2018,mwt:rdpg1} for the case of $1$-dimensional support curves.  As in \cite{mwt:rdpg1}, we consider the case of an unknown support manifold that must be learned from auxiliary data.  Whereas \cite{mwt:rdpg1} studied a specific decision problem (testing the equality of two Fr\'{e}chet means) on a support curve, here we study general decision problems on $d$-dimensional support manifolds.

The semisupervised decision rules studied in this report have a common structure in Steps 1--5, although Step 5 includes three possible embedding strategies.  They vary in Step 6, in the choice of a function $\phi$ that assigns an action to a configuration of $n$ points in $\Re^d$.  This report does not consider the problem of finding an optimal $\phi$, which will (of course) depend on the decision problem.  Instead, for (almost) any $\phi$, we have demonstrated that the performance of $\phi$ when $\M$ is known can be approximated when $\M$ is unknown.
We reiterate that $n$, the size of the sample to which $\phi$ is applied, is fixed.  It is $N$, the size of the auxiliary sample used to learn $\M$, that we allow to increase.  In the limit, we extract the full extent of the $d$-dimensional Euclidean structure in $\M$.  Approximating this structure offers hope that $\phi$ can exploit some benefit from restricted inference, but $n$ may be small and the benefit slight.  We offer no guarantees that the resulting risk will be small.

It is also possible that performance superior to that of $\phi$ can be obtained by a decision rule that exploits non-Euclidean structure.  Such rules might preserve Step 1--4 of our rules, but replace Steps 5--6 with functions that assign actions based in other ways on the dissimilarity function that approximates Riemannian distance on $\M$.  This is a topic for future research.

The centerpiece of our investigation is Section~\ref{dissim}, in which we study the behavior of dissimilarity functions constructed from shortest path distances on localization graphs.  Our analysis extends results in \cite{Bernstein&etal:2000} from the case of points that lie on the manifold to the case of points that uniformly approach the manifold.  In particular, Theorems \ref{thm:SPD0} and \ref{thm:SPDconverge} state conditions that {\em allow}\/ convergence in ratio to the Riemannian distance function.  These conditions have broad implications for designing algorithms that implement Isomap for manifold learning.  For example, it is imperative that $\epsilon_1$ be strictly less than $\epsilon_2$, excluding the popular use of $\epsilon$-neighborhood graphs for localizing data.  We further explore the design of Isomap algorithms in a sequel.

It is important to recognize that, while Theorem~\ref{thm:SPDconverge} establishes the {\em existence}\/ of convergent sequences, it does not construct an algorithm that is guaranteed to converge to the desired limit.  The stated conditions involve parameters of the support manifold, which is unknown.  It is therefore unclear how to determine actual values of $N(j)$, $\epsilon_1(j)$, and $\epsilon_2(j)$ that guarantee the desired convergence.

Finally, the proof of Theorem~\ref{thm:SPDconverge} requires $\delta(j)$, the density of the auxiliary sample obtained via Lemma~\ref{lm:sampling}, to be less than $\sigma_1(j)$, the maximum distance between the true and estimated latent positions.  Theorem~\ref{thm:2inf} guarantees that $\sigma_1(j)$ converges in probability to zero, but the $N(j)$ required to achieve a small $\sigma_1$ with high probability is typically quite large.  Likewise, the $N(j)$ required to achieve a small $\delta < \sigma_1$ with high probability will be quite large.  Thus, large auxiliary samples are required to achieve good approximations.

Despite the caveats expressed above, we believe that the framework and theoretical results reported herein offer substantial promise for developing practical methods for restricted inference on random dot product graphs.  A sequel will describe various examples and applications, and report the results of simulation experiments.

\section*{Appendix A: Manifold Parameters}

Let $\M \subset \Re^q$ denote a $p$-dimensional Riemannian manifold, let $\kappa_1(m)$ denote the first principal curvature at $y \in \M$, and let 
$\bar{\kappa}_1(\M) = \sup \left\{ \kappa_1(y) : y \in \M \right\}$.
If $\M$ is both smooth and compact, then the supremum in $\bar{\kappa}_1(\M)$ is attained and its reciprocal, $r_0(\M)$, is the {\em minimum radius of curvature}\/ of $\M$.

Following \cite{Alexander&etal:1987},
a real number $\sigma$ is a {\em tubular radius}\/ for $\M \subset \Re^q$ if every point in $\Re^q$ at Euclidean distance $\sigma$ or less from $\M$ is the center of a closed ball in $\Re^q$ that meets $\M$ at a single point.
If $\sigma$ is a tubular radius, then the
{\em tubular neighborhood}\/ $\M_\sigma$ that consists of every point in $\Re^q$ at Euclidean distance $\sigma$ or less from $\M$ is a $q$-dimensional Riemannian manifold.  If $\M$ is smooth and compact, then the {\em Tubular Neighborhood Theorem}\/ guarantees the existence of a strictly positive tubular radius.

If $\sigma$ is a tubular radius for $\M$, then so is any nonnegative real number less than $\sigma$.  The supremum of the tubular radii of $\M$, denoted $\bar{\sigma}(\M)$, is the {\em reach}\/ of $\M$.
The magnitude of the reach is controlled locally by the minimum radius of curvature: $\bar{\sigma}(\M) \leq r_0(\M)$.  It is controlled globally by how close $\M$ comes to self-intersection.
In \cite{Bernstein&etal:2000}, that quantity is called the {\em minimum branch separation}\/ and denoted $s_0(\M)$.  If $\M$ is both smooth and compact, then $\bar{\sigma}(\M) = \min \{ r_0(\M), s_0(\M)/2 \}$.

The tubular manifold $\M_\sigma$ has minimum branch separation 
$s_0 \left( \M_\sigma \right) = s_0(\M)-2\sigma$,
but its minimum radius of curvature depends on the relation between $p$ and $q$.  For example, if $p=1$ (i.e., $\M$ is a curve) and $q=2$, then
$r_0 \left( \M_\sigma \right) = r_0(\M)$;
whereas if $p=1$ and $q=3$, then
$r_0 \left( \M_\sigma \right) = \sigma$.
Regardless,
$r_0 \left( \M_\sigma \right) \geq \min \left\{ r_0(\M), \sigma \right\} = \sigma$.
If $\sigma$ is sufficiently small, then 
$r_0 \left( \M_\sigma \right) = \sigma$.

Because $\M$ is a submanifold of the Riemannian manifold $\M_\sigma$, Riemannian distance on $\M_\sigma$ restricted to $\M$ is a distance on $\M$.  The following inequalities relate this induced distance function to Riemannian distance on $\M$.
\begin{lemma}[\cite{Kohan:2020}]
Let $\M \subset \Re^q$ be a smooth compact Riemannian manifold with Riemannian distance function $d_0$ and minimum radius of curvature $r_0$.  Let $\sigma>0$ be a tubular radius of $\M$ and let $d_\sigma$ denote Riemannian distance on $\M_\sigma$.
Then, for every $y_1,y_2 \in \M$,
\[
0 \leq d_0 \left( y_1,y_2 \right) - 
d_\sigma \left( y_1,y_2 \right) \leq \sigma \, \mbox{\tt diam}(\M)/r_0,
\]
where $\mbox{\tt diam}(\M)$ is the extrinsic diameter of $M$, i.e., the diameter of $\M$ measured with respect to Euclidean distance.
\label{lm:tubular}
\end{lemma}
It follows immediately that a sequence of functions $d_{\sigma_k} : \M \times \M \rightarrow \Re$ converges uniformly to $d_0 : \M \times \M \rightarrow \Re$ as $\sigma_k \rightarrow 0$.
To obtain a stronger mode of convergence, we will also rely on the following inequalities.
\begin{lemma}[\cite{Bernstein&etal:2000}, Corollary~4]
Let $\M \subset \Re^q$ be a smooth compact Riemannian manifold with Riemannian distance function $d_0$, minimum radius of curvature $r_0$, and minimum branch separation $s_0$.
If $y_1,y_2 \in \M$ satisfy 
\begin{eqnarray*}
\left\| y_1-y_2 \right\| < s_0 & \mbox{and} &
\left\| y_1-y_2 \right\| \leq \frac{4}{\pi} r_0 \sqrt{6\lambda},
\end{eqnarray*}
then
\[
(1-\lambda) d_0 \left( y_1,y_2 \right) \leq 
\left\| y_1 - y_2 \right\| \leq
d_0 \left( y_1,y_2 \right).
\]
\label{lm:euclid}
\end{lemma}

\section*{Acknowledgments}
Support for this effort was provided by Defense Advanced Research Projects Agency (DARPA) Artificial Intelligence Quantified (AIQ) award number HR00112520026.

\bibliography{$HOME/lib/tex/stat,$HOME/lib/tex/mds,$HOME/lib/tex/math,$HOME/lib/tex/mwt,$HOME/lib/tex/cep,$HOME/lib/tex/bio,$HOME/lib/tex/net}

\begin{thebibliography}{22}
\providecommand{\natexlab}[1]{#1}
\providecommand{\url}[1]{\texttt{#1}}
\expandafter\ifx\csname urlstyle\endcsname\relax
  \providecommand{\doi}[1]{doi: #1}\else
  \providecommand{\doi}{doi: \begingroup \urlstyle{rm}\Url}\fi

\bibitem[Alexander et~al.(1987)Alexander, Bray, and
  Bishop]{Alexander&etal:1987}
S.~Alexander, I.~D. Bray, and R.~L. Bishop.
\newblock The {R}iemannian obstacle problem.
\newblock \emph{Illinois Journal of Mathematics}, 31\penalty0 (1):\penalty0
  167--184, 1987.

\bibitem[Anderson et~al.(1992)Anderson, Wasserman, and
  Faust]{anderson&etal:1992}
C.~J. Anderson, S.~Wasserman, and K.~Faust.
\newblock Building stochastic blockmodels.
\newblock \emph{Social Networks}, 14:\penalty0 137--161, 1992.

\bibitem[Athreya et~al.(2018)Athreya, Fishkind, Levin, Lyzinski, Park, Qin,
  Sussman, Tang, Vogelstein, and Priebe]{AA&etal:2018}
A.~Athreya, D.~E. Fishkind, K.~Levin, V.~Lyzinski, Y.~Park, Y.~Qin, D.~L.
  Sussman, M.~Tang, J.~T. Vogelstein, and C.~E. Priebe.
\newblock Statistical inference on random dot product graphs: A survey.
\newblock \emph{Journal of Machine Learning Research}, 18\penalty0
  (226):\penalty0 1--92, 2018.

\bibitem[Athreya et~al.(2021)Athreya, Tang, Park, and Priebe]{LSM:2018}
A.~Athreya, M.~Tang, Y.~Park, and C.~E. Priebe.
\newblock On estimation and inference in latent structure random graphs.
\newblock \emph{Statistical Science}, 36\penalty0 (1):\penalty0 68--88, 2021.

\bibitem[Bernstein et~al.(2000)Bernstein, de~Silva, Langford, and
  Tenenbaum]{Bernstein&etal:2000}
M.~Bernstein, V.~de~Silva, J.~C. Langford, and J.~B. Tenenbaum.
\newblock Graph approximations to geodesics on embedded manifolds.
\newblock \verb+https://web.mit.edu/cocosci/isomap/BdSLT.pdf+, December 20,
  2000.

\bibitem[Billingsley(1971)]{billingsley:1971}
P.~Billingsley.
\newblock \emph{Weak Convergence of Measures: Applications in Probability}.
\newblock Society for Industrial and Applied Mathematics, Philadelphia, 1971.

\bibitem[Cape et~al.(2019)Cape, Tang, and Priebe]{2toInf:2019}
J.~Cape, M.~Tang, and C.~E. Priebe.
\newblock The two-to-infinity norm and singular subspace geometry with
  applications to high-dimensional statistics.
\newblock \emph{Annals of Statistics}, 47\penalty0 (5):\penalty0 2405--2439,
  2019.

\bibitem[Gilbert(1959)]{Gilbert:1959}
E.~N. Gilbert.
\newblock Random graphs.
\newblock \emph{Annals of Mathematical Statistics}, 30\penalty0 (4):\penalty0
  1141--1144, 1959.

\bibitem[Hoff et~al.(2002)Hoff, Raftery, and Handcock]{hoff&etal:2002}
P.~D. Hoff, A.~E. Raftery, and M.~S. Handcock.
\newblock Latent space approaches to social network analysis.
\newblock \emph{Journal of the American Statistical Association}, 97\penalty0
  (460):\penalty0 1090--1098, 2002.

\bibitem[Holland et~al.(1983)Holland, Laskey, and Leinhardt]{holland&etal:1983}
P.~W. Holland, K.~B. Laskey, and S.~Leinhardt.
\newblock Stochastic blockmodels: First steps.
\newblock \emph{Social Networks}, 5:\penalty0 109--137, 1983.

\bibitem[Kohan(2020)]{Kohan:2020}
M.~Kohan.
\newblock {G}romov {H}ausdorff distance to tubular neighborhood.
\newblock Math{O}verflow, 2020.
\newblock {URL}: \mbox{\tt https://mathoverflow.net/q/369199} (version:
  2020-08-16).

\bibitem[Kruskal(1964)]{kruskal:1964a}
J.~B. Kruskal.
\newblock Multidimensional scaling by optimizing goodness of fit to a nonmetric
  hypothesis.
\newblock \emph{Psychometrika}, 29:\penalty0 1--27, 1964.

\bibitem[Lyzinski et~al.(2017)Lyzinski, Tang, Athreya, Park, and
  Priebe]{VL&etal:2017}
V.~Lyzinski, M.~Tang, A.~Athreya, Y.~Park, and C.~E. Priebe.
\newblock Community detection and classification in hierarchical stochastic
  blockmodels.
\newblock \emph{{IEEE} Transactions on Network Science and Engineering},
  4\penalty0 (1):\penalty0 13--26, 2017.

\bibitem[Rubin-Delanchy(2020)]{PRD:2020}
P.~Rubin-Delanchy.
\newblock Manifold structure in graph embeddings.
\newblock arXiv:2006.05168, 2020.

\bibitem[Rubin-Delanchy et~al.(2022)Rubin-Delanchy, Cape, Tang, and
  Priebe]{GRDPG:2022}
P.~Rubin-Delanchy, J.~Cape, M.~Tang, and C.~E. Priebe.
\newblock A statistical interpretation of spectral embedding: The generalised
  random dot product graph.
\newblock \emph{Journal of the Royal Statistical Society: Series B (Statistical
  Methodology)}, 84\penalty0 (4):\penalty0 1446--1473, 2022.

\bibitem[Tenenbaum et~al.(2000)Tenenbaum, de~Silva, and Langford]{isomap:2000}
J.~B. Tenenbaum, V.~de~Silva, and J.~C. Langford.
\newblock A global geometric framework for nonlinear dimensionality reduction.
\newblock \emph{Science}, 290:\penalty0 2319--2323, 2000.

\bibitem[Trosset and B\"{u}y\"{u}kba{\c{s}}(2020)]{mwt:isomap}
M.~W. Trosset and G.~B\"{u}y\"{u}kba{\c{s}}.
\newblock Rehabilitating {I}somap: {E}uclidean representation of geodesic
  structure.
\newblock arXiv:2006.10858, 2020.

\bibitem[Trosset and Priebe(2024)]{mwt:ContinMDS}
M.~W. Trosset and C.~E. Priebe.
\newblock Continuous multidimensional scaling.
\newblock arXiv:2402.04436, 2024.

\bibitem[Trosset et~al.(2020)Trosset, Gao, Tang, and Priebe]{mwt:rdpg1}
M.~W. Trosset, M.~Gao, M.~Tang, and C.~E. Priebe.
\newblock Learning $1$-dimensional submanifolds for subsequent inference on
  random dot product graphs.
\newblock arXiv:2004.07348, 2020.

\bibitem[Whiteley et~al.(2021)Whiteley, Gray, and
  Rubin-Delanchy]{WhiteleyGrayPRD:2021}
N.~Whiteley, A.~Gray, and P.~Rubin-Delanchy.
\newblock Matrix factorisation and the interpretation of geodesic distance.
\newblock arXiv:2106.01260, 2021.

\bibitem[Whiteley et~al.(2022)Whiteley, Gray, and
  Rubin-Delanchy]{WhiteleyGrayPRD:2022}
N.~Whiteley, A.~Gray, and P.~Rubin-Delanchy.
\newblock Discovering latent topology and geometry in data: A law of large
  dimension.
\newblock arXiv:2208.11665, 2022.

\bibitem[Young and Scheinerman(2007)]{RDPG:2007}
S.~J. Young and E.~R. Scheinerman.
\newblock Random dot product graph models for social networks.
\newblock In A.~Bonato and F.~R.~K. Chung, editors, \emph{Algorithms and Models
  for the Web-Graph}, pages 138--149. Springer, Berlin, Heidelberg, 2007.

\end{thebibliography}

\end{document}